\documentclass{article}

\usepackage{arxiv}

\usepackage[utf8]{inputenc} % allow utf-8 input
\usepackage[T1]{fontenc}    % use 8-bit T1 fonts
\usepackage{hyperref}       % hyperlinks
\usepackage{url}            % simple URL typesetting
\usepackage{booktabs}       % professional-quality tables
\usepackage{amsfonts}       % blackboard math symbols
\usepackage{nicefrac}       % compact symbols for 1/2, etc.
\usepackage{microtype}      % microtypography
\usepackage{lipsum}
\usepackage{graphicx}
\graphicspath{ {./images/} }

\usepackage{adjustbox}
\usepackage{array}
\usepackage{multirow} % Required for multirow
\usepackage{amssymb}
\usepackage{float}
\title{High Performance Space Debris Tracking in Complex Skylight Backgrounds with a Large-Scale Dataset}
\usepackage{amsmath}
\usepackage[table]{xcolor}

\author{
Guohang Zhuang\textsuperscript{1,2}\thanks{Equal contribution.},
Weixi Song\textsuperscript{3}\footnotemark[1],
Jinyang Huang\textsuperscript{1},
Chenwei Yang\textsuperscript{4},
Wanli Ouyang\textsuperscript{2,5},
Yan Lu\textsuperscript{2,5}\thanks{Corresponding author.} \\
\textsuperscript{1}Hefei University of Technology,
\textsuperscript{2}Shanghai Artificial Intelligence Laboratory, \\
\textsuperscript{3}Zhejiang University,
\textsuperscript{4}Polar Research Institute of China, \\
\textsuperscript{5}The Chinese University of Hong Kong \\
}

\begin{document}
\maketitle
\begin{abstract}
With the rapid development of space exploration, space debris has attracted more attention due to its potential extreme threat, leading to the need for real-time and accurate debris tracking. However, existing methods are mainly based on traditional signal processing, which cannot effectively process the complex background and dense space debris.
In this paper, we propose a deep learning-based Space Debris Tracking Network~(SDT-Net) to achieve highly accurate debris tracking. SDT-Net effectively represents the feature of debris, enhancing the efficiency and stability of end-to-end model learning. To train and evaluate this model effectively, we also produce a large-scale dataset Space Debris Tracking Dataset~(SDTD) by a novel observation-based data simulation scheme. SDTD contains 18,040 video sequences with a total of 62,562 frames and covers 250,000 synthetic space debris. Extensive experiments validate the effectiveness of our model and the challenging of our dataset. Furthermore, we test our model on real data from the Antarctic Station, achieving a MOTA score of 73.2\%, which demonstrates its strong transferability to real-world scenarios. Our dataset and code will be released soon.
\end{abstract}

\begin{figure*}[t!]
\centering %表示居中
\includegraphics[width=1.0\columnwidth]{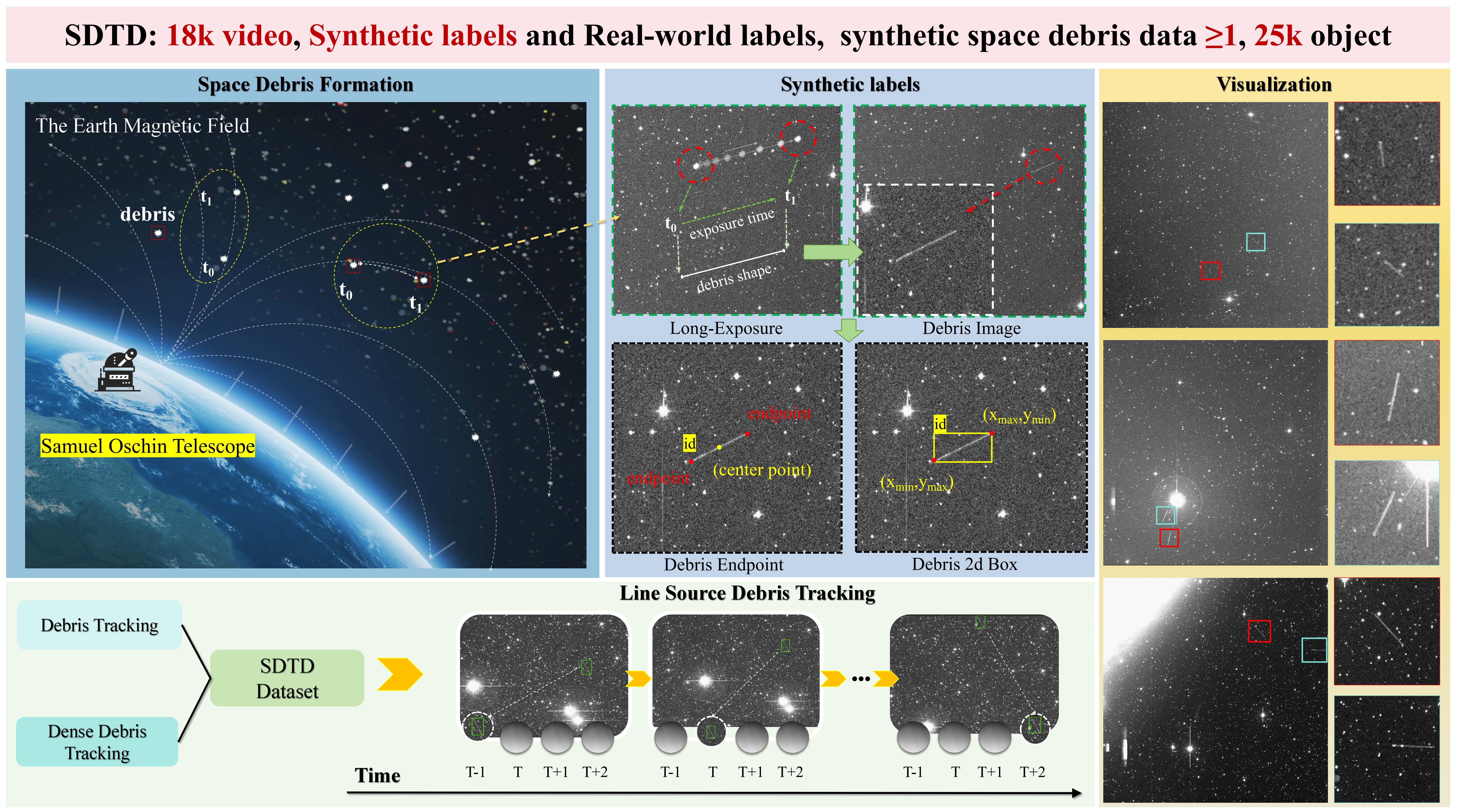}
\caption{Illustration of the space debris formation and tracking pipeline using synthetic labels. It includes the simulation of space debris within Earth's magnetic field, the generation of synthetic labels such as start (st), end (ed), and center (cen) points, and bounding boxes for debris in image frames. Additionally, the process of segment source debris tracking across multiple frames is shown, with visualization examples of debris tracking and dense debris tracking using the SDTD dataset.}
%图片的名称
\label{fig1}
\vspace{-8pt}
%图片的标签，用于文章中的引用，注意到标签的数字与实际文章显示的数字可能不同
\end{figure*}

% keywords can be removed
%\keywords{First keyword \and Second keyword \and More}

\section{Introduction}
During the past decade, global space activities have exponentially increased, which has raised the potential risk of collisions between debris and spacecraft in near-Earth space~\cite{katz2024can,steindorfer2025space}. This has led to significant economic losses, drawing increasing attention to the issue of space debris. The Space Debris Tracking (SDT) is to detect and predict the trajectories of space debris, reducing the collision risks and promoting space industry development. This makes monitoring debris extremely urgent and critical for ensuring the safety of space operations.
% 密集目标和复杂背景

SDT is an essential task that includes detecting and predicting the trajectories of space debris. It is a computer vision problem, specifically an object tracking task in astronomical videos, where the goal is to identify and follow space debris across continuous frames for a video. However, traditional methods typically focus on object debris detection rather than tracking. These approaches often rely on techniques such as template-matching methods~\cite{LIU2012380, murphy2017space} and morphological operator methods~\cite{wei2018real, jiang2022automatic}. While these methods are capable of detecting debris in individual images, they struggle to associate objects across consecutive frames, making them less suitable for space debris monitoring.

Artificial intelligence~(AI) has achieved significant success in fields such as object detection~\cite{zou2023object} and object tracking~\cite{javed2022visual}, due to its exceptional ability to process large-scale datasets. Traditional methods have limitations in space debris detection and cannot effectively handle long-term tracking tasks. AI models can make up for this deficiency through efficient data processing and learning methods. However, AI models rely on large-scale, high-quality datasets while astronomy data are extremely limited. So, there is no an effective AI model that could meet the requirements of debris tracking mission for popular telescopes like the Solar Data Telescope~(LST)~\cite{li2019lyman} and Wide-Field Survey Telescope~(WFST)~\cite{wang2023science}.

% \begin{figure*}[t!]
% \centering %表示居中
% \includegraphics[width=1.0\columnwidth]{fig1.jpg}
% \caption{Illustration of the space debris formation and tracking pipeline using synthetic labels. It includes the simulation of space debris within Earth's magnetic field, the generation of synthetic labels such as start (st), end (ed), and center (cen) points, and bounding boxes for debris in image frames. Additionally, the process of segment source debris tracking across multiple frames is shown, with visualization examples of debris tracking and dense debris tracking using the SDTD dataset.}
% %图片的名称
% \label{fig1}
% \vspace{-8pt}
% %图片的标签，用于文章中的引用，注意到标签的数字与实际文章显示的数字可能不同
% \end{figure*}

% \includegraphics[width=1.9\columnwidth]{Fig/fig1_9.jpg} % 双栏情况下宽度为 columnwidth
% \captionof{figure}{Illustration of the space debris formation and tracking pipeline using synthetic labels. It includes the simulation of space debris within Earth's magnetic field, the generation of synthetic labels such as start (st), end (ed), and center (cen) points, and bounding boxes for debris in image frames. Additionally, the process of segment source debris tracking across multiple frames is shown, with visualization examples of debris tracking and dense debris tracking using the SDTD dataset.}
% \label{fig1}

To address the challenge of limited data in astronomy, we follow the idea of observation-based data simulation in the astronomical field~\cite{lin2021new,peterson2015simulation,perrin2014updated}. The observation-based data simulation is widely proven to lead to realistic data and reliable scientific conclusions, providing a chance for us to design a method for data simulating space debris. Due to its success, we construct a Space Debris Tracking Dataset~(SDTD), which is the first publicly available dataset designed for space debris tracking. Fig~\ref{fig1} shows the formation principle, type, examples, and corresponding SDT tasks of space debris. Specifically, Tab~\ref{t1} highlights the key differences of SDTD compared to existing datasets. SDTD consists of 18,040 synthetic videos, totaling 62,562 frames and 25,000 annotated debris instances.

% \textcolor{red}{Tab~\ref{t1} highlights the key differences of SDTD compared to existing datasets. Specifically, as shown in Fig~\ref{fig1}, SDTD consists of 18,040 synthetic videos, totaling 62,562 frames and 25,000 annotated debris instances. }

In order to apply AI methods to the SDT task, we propose a simple yet effective deep learning method, namely the Space Debris Tracking Network~(SDT-Net). We introduce a Region-of-Interest Feature Enhancement (RoIFE) module to highlight debris features, aiding in more accurate detection. A detection module is then used to localize the debris by predicting their positions in the image. After localization, a debris offset module is applied to perform cross-frame data association, generating tracking trajectories for each debris.

In summary, the key contributions of this paper are as follows:
% 强调价值，强调这个的意义

1. We propose a space debris simulation method. Based on this method, we construct SDTD, the first benchmark dataset for SDT. It contains 18,040 video data with complex backgrounds and diverse debris targets, and the data is close to the real space debris situation.

2. We introduce the Space Debris Tracking Network (SDT-Net) to improve space debris tracking. By integrating a RoIFE module, a detection module, and a tracking module, we effectively enhance debris representations, leading to more accurate and reliable tracking performance.
% segmentation base feature enhancement

3. Evaluation on the constructed SDTD dataset shows that the proposed SDT-Net achieves state-of-the-art performance in terms of space debris tracking on the test set overall.

\begin{table*}[htbp!]
\centering
\caption{Summary of Space Debris Target Datasets.}
\label{t1}
\small % 使用较小的字体
\setlength{\tabcolsep}{4pt} % 调整列间距
\renewcommand{\arraystretch}{1.2} % 调整行间距
\resizebox{\textwidth}{!}{ % 缩小表格宽度
\begin{tabular}{@{}lccccccccc@{}}
\toprule
\textbf{Dataset} & \textbf{Type} & \textbf{Size} & \textbf{\#Video.} & \textbf{\#Det.} & \textbf{\#Seg.} & \textbf{\#Track.}  & \textbf{Target Variations} & \textbf{Debris Number} \\ 
\midrule
Lin et al.~\cite{lin2021new} & Real & 83 images & 7 video & $\checkmark$ & $\times$ & $\times$  & One & One \\

% \cite{yao2022adaptive} & Real & 100 images  & $\checkmark$ & $\times$ & $\times$ & Point target & Two & One \\

SDebrisNet \cite{tao2023sdebrisnet} & Real & 1551 images & $\times$  & $\checkmark$ & $\times$ & $\times$  & Three & One \\

Jiang et al.~\cite{jiang2022automatic} & Real & 200 images & $\times$ & $\checkmark$ & $\times$ & $\times$  & One & One \\

% \cite{liu2020space} & Real & 111 images   & $\checkmark$ & $\times$ & $\times$ & Point target & One & One \\

AstroStripeSet \cite{zhu2024sstd} & Real & 1500 images & $\times$ & $\checkmark$ & $\checkmark$ & $\times$  & Four & One \\

\rowcolor[gray]{0.9}
SDTD & Synthetic \& Real & 65,562 images  & 18,040 video & $\checkmark$ & $\checkmark$ &\textbf{ $\checkmark$}  & Five & One to Five \\
\bottomrule
\vspace{-10pt}
\end{tabular}}
\end{table*}

\section{Related Work}
\label{sec:Related Work}

\subsection{Object Detection and Tracking}
Object detection and tracking involves detecting objects, assigning a unique ID to each detection, and then tracking these objects across frames in a video while maintaining the ID assignments. This task has been a central focus in computer vision and has found applications in various fields, such as autonomous driving~\cite{geiger2012we}. 

In the context of tracking, several methods have been proposed over the years. For instance, SORT~\cite{bewley2016simple} conducted track association by combining Kalman Filter~\cite{welch1995introduction} and Hungarian
algorithm~\cite{kuhn1955hungarian}. DeepSORT~\cite{wojke2017simple} introduced an extra cosine distance and compute the appearance similarity for track association. CenterTrack~\cite{zhou2020tracking} is a center-based multi-object tracking method that extends the detection framework of CenterNet \cite{zhou2019objects} to achieve efficient real-time tracking. ByteTrack~\cite{zhang2022bytetrack} exploits low-score detection boxes by first matching high-confidence detections and then associating them with low-score detections.

\subsection{Space Debris Detection}
% \subsection{Detection and Tracking of Space Debris}
%把space debris detection或者tracking的定义再讲一下
Space Debris Detection is a task of predicting space debris location from a single image. Jiang et al.~\cite{jiang2022space} applied enhanced median filtering to remove noise and used an improved Hough transform to detect targets. Guo et al.~\cite{guo2025enhanced} proposed an enhanced YOLOv8-based method for space debris detection, achieving improved accuracy and speed. Jia et al.~\cite{jia2020detection} proposed a deep neural network-based framework for detecting and classifying astronomical targets. Specifically, they utilized Faster R-CNN~\cite{ren2015faster} with a modified ResNet-50~\cite{he2016deep} backbone to improve the detection of low signal-to-noise targets. Previous studies have primarily focused on detection tasks, limiting their applications in space debris. In this paper, we extend the detection to tracking, playing a key role in advancing space exploration.

\section{Benchmark Dataset}
\label{sec:Benchmark}

\begin{figure*}[t!]
\centering %表示居中
\includegraphics[width=1.0\columnwidth]{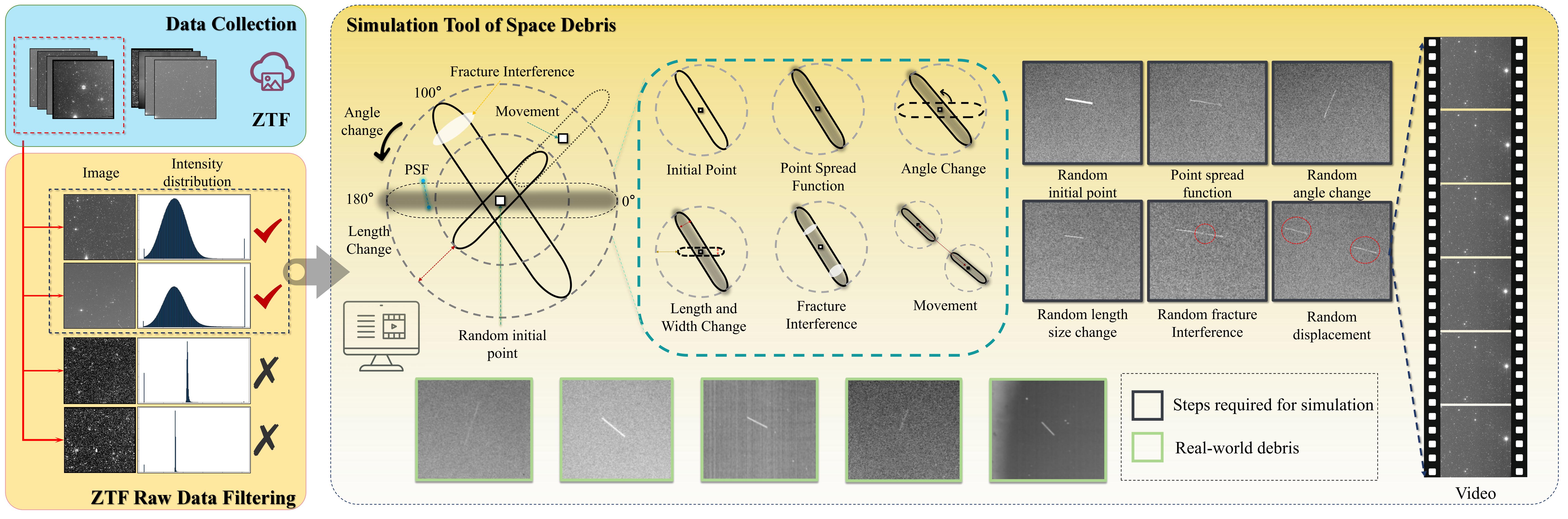}
\caption{Overview of the space debris simulation pipeline. Step 1: Data Collection. Step 2: ZTF Raw Data Filtering. Step 3: Simulation Tool for Space Debris. Various parameters, including random initial points, angle change, length change, point spread function (PSF), fracture interference, and displacement, are applied to simulate space debris. The process generates synthetic video sequences and shows the real-world debris for reference.}
%图片的名称
\label{fig3}
\vspace{-8pt}
%图片的标签，用于文章中的引用，注意到标签的数字与实际文章显示的数字可能不同
\end{figure*}

In this paper, we utilize astronomical images collected by the Zwicky Transient Facility (ZTF)~\cite{bellm2014zwicky} to simulate space debris. We propose a simulation tool to generate space debris, which provides sufficient data for constructing the SDTD dataset. Under long-exposure conditions in telescopic observations, debris often appears as a line source in the images. So, our simulation target is to generate realistic long-exposure line sources. We develop a pipeline that includes data pre-processing, line source debris video generation, and post processing. Fig~\ref{fig3} and the following paragraphs provide a detailed description of this process.

\subsection{Data Collection and Pre-Processing}

\subsubsection{Data Collection}
The ZTF is an optical survey project designed to monitor astronomical transient phenomena. It operates on the 48-inch Samuel Oschin Telescope at Palomar Observatory, using a 47-square-degree, 600-megapixel CCD camera for high-cadence, wide-field imaging. Its capability to capture large-scale, high-resolution astronomical images makes it suitable for simulating space debris.

In order to obtain enough data for simulation, we collected 17,235 ZTF astronomical observation data from June 2018 to June 2022. These data contain complex skylight background information, including atmospheric scattered light, moonlight, and human factors such as light pollution. Based on these richer astronomical observation data, we conduct simulations and generated debris videos.

Some regions of the images contain outliers, appearing as zeros or extremely high values, which leads to numerical overflow issues. To ensure the quality of our dataset, we carefully examine the dataset, remove 1,195 unusable images, and retain 16,040 high-quality images. 

\subsubsection{Data Pre-Processing}
Unlike natural images, astronomical images typically contain larger pixel values. Therefore, a data processing step is necessary to transform these values into the standard range of grayscale image pixels. To achieve this, we apply the ZScale method \cite{payne2003astronomical} to map image pixel values to a reasonable range. The ZScale method first calculates the median $z_{median}$ and standard deviation $z_{std}$ of the image pixels. Then uses these two values to estimate a range $[z_1,z_2]$:
\begin{equation}
z_1 = z_{median} - k\times z_{std}, 
z_2 = z_{median} + k\times z_{std},
\end{equation}
where $z_1$ and $z_2$ represent the lower and upper bounds of the pixel intensity range, respectively. The hyperparameter $k$ is assigned a value of 2.5. Then, we perform a linear stretch on the grayscale values of the image, mapping the range $[z_1, z_2]$ to $[0, 255]$: 

\begin{equation}
I_{scaled}(x,y) = 255 \times \frac{I(x, y) - z_1}{z_2 - z_1}.
\end{equation}

Finally, pixels with grayscale values below $z_1$ or above $z_2$ are set to 0 or 255:
\begin{equation}
I_{\text{final}}(x, y) =
\begin{cases}
\max(I_{\text{scale}}(x, y), 0), & I_{\text{scale}}(x, y) < 0 \\
\min(I_{\text{scale}}(x, y), 255), & I_{\text{scale}}(x, y) > 0
\end{cases}
\end{equation}

% \begin{equation}
% I_{final}(x,y) = min((max(I_{scaled}(x,y),0),255).
% \end{equation}

By enhancing the contrast and detail of astronomical images, the ZScale method ensures a reliable basis for the next simulations.

After pre-processing the data, we conduct simulations to generate space debris, providing sufficient data for constructing the SDTD dataset. Details of the simulation process are described in Section~\ref{s4}.

\subsection{Simulation of Space Debris}\label{s4}
For a single astronomical image, the Space Debris Simulation generates a video with $T$ frames of line source debris, simulating the movement of debris across each frame.

\subsubsection{Line Source Debris Video Generation}

The goal of the space debris simulation is to generate a sequence video with multiple line source debris. 
For each debris (set $k$-th as an example), we represent it as a parameter sequence $\{S_k^1,S_k^2,S_k^3,...S_k^T\}$ where $T$ is the total frame number. $S_k^t$ is the debris parameters for the $t$-th frame, which is set as follows:
\begin{equation}
S_k^t = \{\underbrace{x^t_k,y^t_k}_{\text{center}}, \underbrace{l_k, w_k}_{\text{dimension}}, \underbrace{\theta_k, v_k}_{\text{velocity}}\},
\end{equation}
where $(x^t_k,y^t_k)$ represents center point coordinates of the line source at time $t$. $l_k$, $w_k$, $\theta_k$ and $v_k$ are line length, line width, velocity angle and velocity magnitude, respectively.

With this parametrized line source, we randomly sample a value $K$ as the total line source number and then for the $k$-th line source, we first generate its $S^1_k$ by randomly sampling its 6 parameters. The generated $S^1_k$ provide the initial center point locations, defined dimension, and a fixed velocity of the debris. 
So, we utilize these parameters to produce subsequent parameters $S^2_k\sim S^T_k$.

Take the second frame~$(t=2)$ as an example, we calculate the parameters for $S^2_k$. The center point coordinates $(x^2_k,y^2_k)$ is calculated by combining velocity and initial center as follows: 
\begin{equation}
\begin{aligned}
x^2_k=x^1_l+v_k \cdot \cos(\theta_k) \cdot \Delta t,\\
y^2_k=y^1_k+v_k \cdot \sin(\theta_k) \cdot \Delta t,
\end{aligned}
\end{equation}
where $\Delta t$ represents the time interval between consecutive frames, which is set to 1 here for simplification. Thus, the parameters for the second frame are represented as $S_k^2 =\{x^2_k,y^2_k, l_k, w_k, \theta_k, v_k\}$.
The same procedure is repeated for each subsequent frame $t=\{3,4,5,...,T\}$. This process continues frame by frame, until all parameters $\{S_k^1,S_k^2,S_k^3,...S_k^T\}$ are generated.

We proceed to generate the corresponding line source in the image. For each frame $t$, we take the parameters $S^t_k=\{x^t_k,y^t_k, l_k, w_k, \theta_k, v_k\}$ and follow three steps to generate the line source. With these, at the center point coordinates $(x^t_k,y^t_k)$, we draw a rectangle with dimensions, $l_k$ and $w_k$, and angle $\theta_k$. Then, we apply random brightness noise to the rectangle. By traversing $k$, we draw entire $K$ objects and if the rectangle is out of the image, we would neglect it. 

At last, using the parameters $\{S_k^1,S_k^2,S_k^3,...S_k^T\}$, we generate the corresponding images $\{I^1,I^2,I^3,...,I^T\}$, resulting in a video that captures the movement trajectory of the debris.

\subsubsection{Post Processing for Realistic}
 % Point Diffusion Function Process
Since the real line source debris conforms to the Point Spread Function~(PSF) characteristics~\cite{kouprianov2008distinguishing}, we perform point diffusion function processing on the preliminary line source debris. Among them, PSF is expressed as follows:
\begin{equation}
I^t_k(x,y)=\frac{S}{2{\pi}{\delta^2_{psf}}} exp(-\frac{(x-x_{c})^2+(y-y_{c}^2)}{2\delta^2_{psf}} ),
\end{equation}
where ${I^t_k(x,y)}$ is the pixel value of the line source debris, $(x_c,y_c)$ represents the center point coordinates of the line source debris, ${S}$ is the scale factor, and ${\delta_{psf}}$ is the diffusion standard deviation of the imaging system PSF. Additionally, we apply random truncation to the line source to simulate more realistic conditions. This operation enables us to simulate realistic line source debris.

\begin{figure*}
\centering %表示居中
\includegraphics[width=1.0\columnwidth]{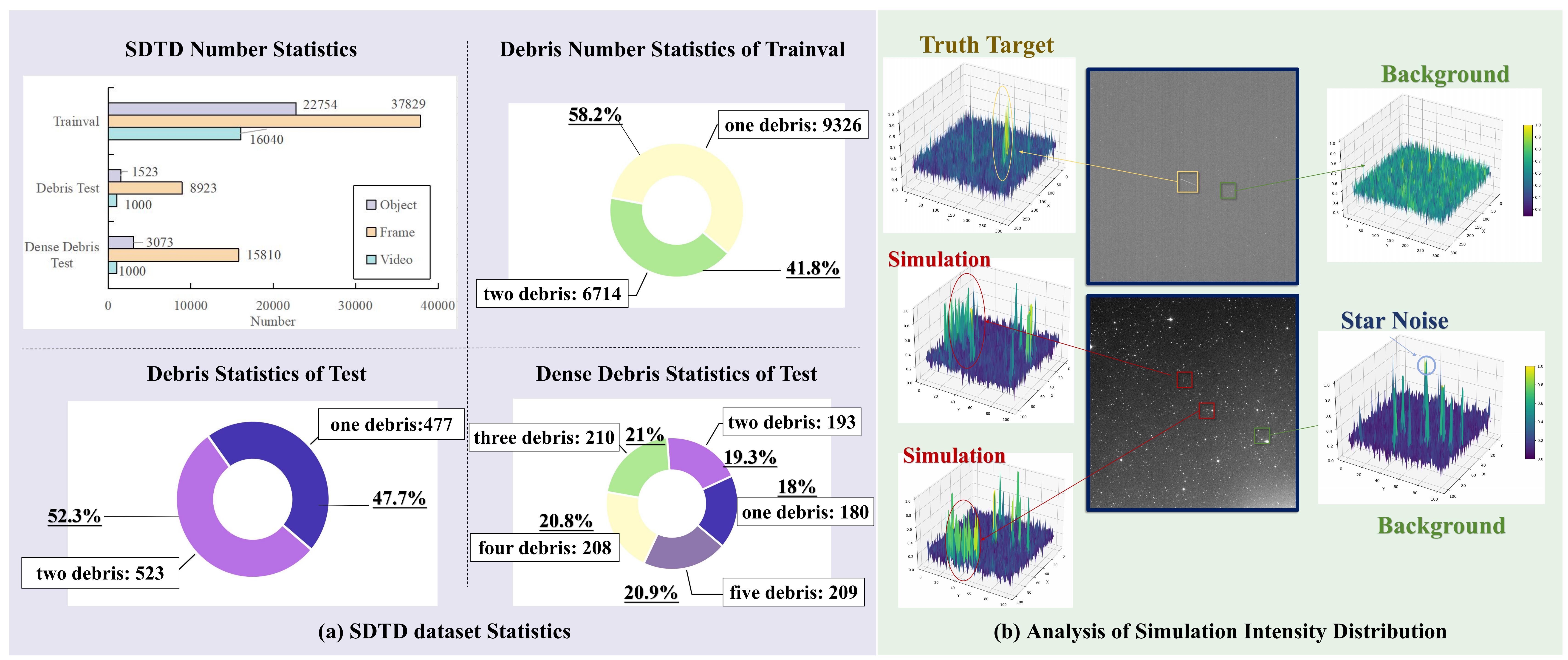}
% [height=4.5cm]表示高度
%[width=9.5cm]表示宽度
%{111.eps}表示eps格式的图片，名为111
\caption{Overview of SDTD dataset statistics and analysis of debris simulation intensity distribution. (a) The left shows the number statistics for training, test, and dense debris categories in the SDTD dataset. The pie charts represent the distribution of debris across different objects in both the training and test sets. (b) The right displays the analysis of simulation intensity distribution, comparing the target, simulation, background, and star noise. It also shows the intensity distribution for debris and background noise.}
%图片的名称
\label{fig2}
\vspace{-8pt}
%图片的标签，用于文章中的引用，注意到标签的数字与实际文章显示的数字可能不同
\end{figure*}

\subsection{Statistics and Visualization}
\subsubsection{Dataset Statistics}
After the simulation, we obtain 18,040 videos with a total of 62,562 frames. As shown in Fig.~\ref{fig2}(a), to evaluate model performance in complex environments, we select 1,000 images from the 16,040 dataset to create two test sets: the debris test set and the dense debris test set. These two test sets allow for a comprehensive evaluation of model robustness under different conditions, including dense debris.

\subsubsection{Debris Type}

To facilitate comprehensive evaluation of model performance at different levels of complexity, we classify the space debris in our dataset into two categories: debris type and dense debris type. This distinction allows us to systematically evaluate tracking robustness in both sparse and complex observation environments.

Debris type refers to space objects that appear singly or sparsely in an observation frame. These objects are separated from each other with limited spatial or temporal overlap, and are therefore relatively easy to detect and track. In contrast, dense debris type represents scenes with high target density, where multiple debris objects (usually more than three) appear within a single frame or between adjacent frames. These instances typically exhibit significant spatial overlap and motion correlation, which increases the difficulty of identity assignment and temporal association.

\subsubsection{Simulation Visualization}

Line source debris in the image generally exhibits high-intensity values. To better compare the effect of simulated line source debris, we visualize the intensity distribution of the image. Fig~\ref{fig2}(b) shows the intensity distribution analysis of the target image. On the left side of the image, we visualized the intensity distribution of different regions in three dimensions. In the yellow box area, we show the intensity distribution of the real target, which shows obvious concentrated peaks, representing the intensity characteristics of the real celestial body. These peaks indicate the concentrated distribution area of the target signal.

In contrast, the red box area shows the target distribution generated by simulation, and the intensity distribution of the simulated target shows a similar trend. In addition, the green box area in the image shows the intensity distribution of the background noise. In the simulated image, the background noise (such as star noise) shows multiple peaks, indicating that the distribution of the noise is more dispersed, and strong interference signals appear in some areas. By comparing these distributions, we can verify the effectiveness of the simulation and confirm that the simulated data can still accurately retain the basic structural characteristics of the target under the influence of background noise.

\section{Method}
\subsection{Overview}
\begin{figure*}[t!]
\centering %表示居中
\includegraphics[width=1.0\columnwidth]{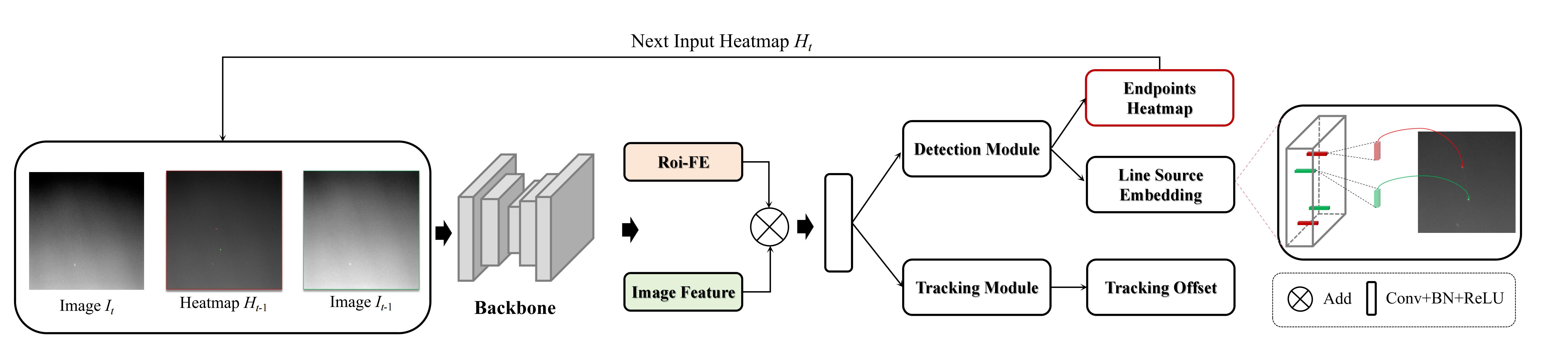}
% [height=4.5cm]表示高度
%[width=9.5cm]表示宽度
%{111.eps}表示eps格式的图片，名为111
\caption{The overall pipeline of our SDT-Net. The inputs are the current frame $I_t$, the previous frame $I_{t-1}$ and heatmap $H_{t-1}$. The outputs include
amodal endpoints heatmap, line source embedding, and tracking offset.}
%图片的名称
\vspace{-8pt}
\label{fig5}
\end{figure*}

The entire pipeline of our Space Debris Tracking Network (SDT-Net) method is shown in Fig.~\ref{fig5}, which is a CenterTrack-style~\cite{zhou2020tracking} tracking method.
Due to the debris long-exposure as linear, we treat the debris tracking as tracking for linear segments.  For each object, we follow the tracking-by-detection idea. Our model detects two endpoints of each line and predicts velocity for cross-frame object association to generate trajectory.

Specifically, a backbone extract base features for each frame, we establish a Region-of-Interest Feature Enhancement~(RoI-FE) module to highlight the debris object cues in feature maps. Then, a novel design detection module consists of an endpoint heatmap head and a line source embedding head that collaborate to localize two endpoints of each debris. After that, a tracking head predicts velocity for each object by computing endpoint offsets across frames.

\subsection{Backbone}
For the $t$-th frame, the model takes three components as inputs, the image of the current frame, the previous frame and its predicted endpoints representing as heatmaps. Each input is passed through a corresponding input-dependent module containing a convolutional layer, a batch normalization layer, and a ReLU activation function. After this, the three feature maps are added together, resulting in a unified feature map $F_{input}\in\mathcal{R}^{W_{in}\times H_{in}\times 3}$. The feature map $F_{input}$ is then passed through our backbone, a DLA model~\cite{yu2018deep}, which consists of an encoder and a decoder. The encoder outputs a series of feature maps at different resolutions, denoted as $\{F_1,F_2,F_3,F_4\}$. 
The decoder, an IDA up-sampler~\cite{yu2018deep}, takes these feature maps to gradually generate upsampled feature maps and we take the highest resolution one as our backbone feature $F_b\in\mathcal{R}^{W\times H\times C}$. 

\subsection{Region-of-Interest Feature Enhancement}

The Region-of-Interest Feature Enhancement module first computes a debris mask.
%, which indicates the debris-related regions. 
% The output debris mask is then multiplied by $F$ to obtain the enhanced feature.
Specifically, we introduce another IDA up-sampler, sharing the same structure with the backbone decoder but having independent parameters, as our segmentation head. Its output channels are set to 1 to obtain a debris mask $\hat M\in\mathcal{R}^{W\times H\times 1}$ based on encoder features. The output mask is multiplied by $F_b$ as follows:
\begin{equation} 
F_{\text{en}} = \hat M \odot F_b
\end{equation}
to acquire the enhanced feature $F_{en}\in\mathcal{R}^{W\times H\times C}$. $F_{en}$ highlights debris features and removes background noise.

To ensure that the obtained mask $\hat S$ accurately indicates the desired debris region, we use a Segmentation Loss to train the segmentation head. The mask $S$ is supervised by a target mask $S$ with the following loss function:
\begin{equation}
% \small
\mathcal{L}_{\text{seg}} = -\frac{1}{WH} \sum_{i=1,j=1}^{W,H} 
\bigg[
M_{ij} \log(\hat{M}_{ij}) + (1 - M_{ij}) \log(1 - \hat{M}_{ij})
\bigg],
\end{equation}
where \(W, H\) are image dimensions, \(M_{ij}\) is the ground truth, and $\hat{M}_{ij}$ is the predicted probability. The obtained $F_{en}$ is then passed through two modules: a Debris Detection module and a Debris Offset module.

\subsection{Debris Detection Module}

In each frame, we detect each line source by detecting its two endpoints. 
Two endpoints of each line source must be one on the left and one on the right. 
So, we design our model to produce a left-endpoint heatmap and a right-endpoint heatmap. The left-endpoint heatmap indicates left endpoints of all line sources, where the right one reflects right points.
And then, we establish an endpoint embedding module to match different left and right endpoints belonging to a same line source together.

\subsubsection{Endpoints Heatmap}

The endpoints heatmap head is a convolutional layer that takes $F_{en}$ as input and feedback heatmaps $\hat H\in\mathcal{R}^{W\times H\times 2}$, which has two channels that one for the left and one for the right endpoint. 
Each pixel in the heatmap represents the likelihood of the position being a line source endpoint. 

Follow existing detection heatmap training pipeline, we utilize the focal loss to train the heatmap head as follows:
\begin{equation}
\mathcal{L}_{\text{hm}} = \frac{1}{N_{\text{pos}}} \sum_{whc} \text{Focal}(H, \hat{H}),
\end{equation}
where $\hat H$ and $H$ are the predicted and the ground-truth heatmaps. $w,h,c$ means the spatial and channel index. 
The ground-truth heatmaps $H$ are generated by drawing Gaussian kernels at the positions of each endpoint.

\subsubsection{Line Source Embedding}

The line source embedding predicts features for each endpoint. 
%The purpose of this is to ensure that each endpoint matches with the corresponding line source. 
Specifically, we predict a left endpoint line source embedding maps $\hat E_{l}\in\mathcal{R}^{W\times H\times C_2}$ and a right endpoint line source embedding maps $\hat E_{r}\in\mathcal{R}^{W\times H\times C_2}$. 
During inference, with the predicted endpoints heatmaps, we first derive local-maximum points for each heatmap, resulting in $N$ left endpoints and $M$ right endpoints. Then we utilize there points to take corresponding embeddings from $\hat E_{l}$ and $\hat E_{r}$, respectively, denoted as $\{e^1_l, e^2_l,...,e^N_l\}$ and $\{e^1_r, e^2_r,...,e^M_r\}$. 
Based on these, we would compute an embedding similarity matrix $S^{emb}\in\mathcal{R}^{N\times M}$ where $S^{emb}_{i,j}$ is computed as follows:
\begin{equation}
S^{emb}_{i,j} = |(e)_l^{i} - (e)_r^{j}|.
\end{equation}
$S^{emb}_{i,j}$ reflects similarity of $i$-th left point and $j$-th right point. With this, we provide a simple yet effective scheme, that each left point matches with its nearest neighbors right point to organize an endpoint pair as a detected object. 

%To ensure the aforementioned functions of embedding, during training, we enforce the line source embeddings of endpoints from the same line source to be close. 
Inspired by CornerNet~\cite{law2018cornernet}, we design the loss function of our embedding as follows: 
\begin{equation}
\mathcal{L}_{\text{same}} = \frac{1}{K} \sum_{k=1}^{K} \left[ (e_l^k - e^k_c)^2 + (e_r^k - e^k_c)^2 \right],
\end{equation}
where \(e^k_l\) and \(e^k_r\) are endpoint features, \(e^k_c\) is the mean of \(e^k_l\) and \(e^k_r\), utilized to push \(e^k_l\) and \(e^k_r\) as close as possible. Meanwhile, we introduce $\mathcal{L}_{\text{diff}}$ to separate the features of different pairs as:
\begin{equation}
\mathcal{L}_{\text{diff}} = \frac{1}{K(K-1)} \sum_{k=1}^{K} \sum_{\substack{j=1 \\ j \neq k}}^{K} \max \left( 0, \Delta - |e^k - e^j| \right),
\end{equation}
where \(\Delta\) is the minimum feature distance threshold. In our experiments, we set \(\Delta\) to 1. This term facilities to push center features belonging to different objects far away.

\subsection{Debris Offset Module}

% a pair of 
We construct a Debris Offset module. For the $t$-th frame, this module predicts two offset maps, $\hat O_l^t\in\mathcal{R}^{W\times H\times 2}$ and $\hat O_r^t\in\mathcal{R}^{W\times H\times 2}$ for left and right endpoints, respectively. 

During inference, the offset plays a key role in data association. Assume that the detection head provides $N$ detected objects, we could index $N$ offsets as: $\{(o_l^t)_1,(o_l^t)_2,(o_l^t)_2...,(o_l^t)_N\}$ and $\{(o_r^t)_1,(o_r^t)_2,(o_r^t)_2...,(o_r^t)_N\}$. They indicate object endpoint offsets between the current frame and the previous $t-1$ frame. With these predicted offsets, for each detected object, we could derive endpoint locations at the previous frame as follows:
\begin{equation}
\text{left: }(c')_l^{t-1} = c_l^t - o_l^t,\ \text{right: }(c')_r^{t-1} = c_r^t - o_r^t,
\end{equation}
where $c'$ is the derived endpoint coordinate and $c$ is the detected one. These derived endpoints imply the potential previous object locations. With these, we could match objects across frames by measuring the location consistency between the derived endpoints and the detected object endpoints. If the previous frame has $M$ detected objects, we would compute a similarity matrix $S^{obj}\in\mathcal{R}^{N\times M}$ where $S^{obj}_{i,j}$ is computed as follows:
\begin{equation}
S^{obj}_{i,j} = |(c')_l^{t-1} - (c)_l^{t-1}| + |(c')_r^{t-1} - (c)_r^{t-1}|.
\end{equation}
$S^{obj}_{i,j}$ reflects similarity of $i$-th object in the $t$-th frame and $j$-th object in the $t-1$ frame. Similar to the embedding module, we also utilize nearest neighbor as a matching scheme that the object in the current frame associates with its nearest neighbors in the previous frame. 
 
To achieve the aforementioned function, during the training phase, we minimize the difference between the ground-truth offset and the predicted offset as follows:
\begin{equation}
\mathcal{L}_{\text{off}} = \frac{1}{K} \sum_{i=1}^{K} \left| (o^t)_i - [(c^t)_i - (c^{t-1})_i] \right|,
\end{equation}

\subsection{Loss function}

Finally, the loss function $\mathcal{L}$ is defined by the weighted summation of all loss terms,
\begin{equation}
% \small
\mathcal{L} = \lambda_{\text{seg}}  \mathcal{L}_{\text{seg}} + \lambda_{\text{hm}}  \mathcal{L}_{\text{hm}} + \lambda_{\text{emb}}  (\mathcal{L}_{\text{same}} + \mathcal{L}_{\text{diff}}) + \lambda_{\text{off}} \mathcal{L}_{\text{off}},
\end{equation}
where $\lambda_{\text{seg}}$, $\lambda_{\text{hm}}$, $\lambda_{\text{emb}}$ and $\lambda_{\text{off}}$ represent the weights for $\mathcal{L}_{\text{seg}}$, $\mathcal{L}_{\text{hm}}$, $(\mathcal{L}_{\text{same}} + \mathcal{L}_{\text{diff}})$ and $\mathcal{L}_{\text{off}}$.

\section{Experiments}
\label{sec:formatting}

\begin{table*}[htbp]
\centering
\small
\caption{Comparison of trackers on Debris Test and Dense Debris Test. The table compares the performance of various tracking algorithms in two distinct scenarios. The \textcolor{red}{\textbf{best}} and \textcolor{blue}{\textbf{second$-$best}} figures are in color and $n$ represents the number of debris.}
\label{tab3}
\resizebox{\textwidth}{!}{
\begin{tabular}{lc*{12}{c}}
\toprule
\multirow{2}{*}{\textbf{Tracker}} & \multicolumn{6}{c}{\textbf{Debris($n \boldsymbol{\leq} 2$)}} & \multicolumn{6}{c}{\textbf{Dense Debris($1 \boldsymbol{\leq} n \boldsymbol{\leq} 5$)}} \\
\cmidrule(lr){2-7} \cmidrule(lr){8-13}
& IDF1↑ & MOTA↑ & IDS↓ & HOTA↑ & DetA↑ & AssA↑ 
& IDF1↑ & MOTA↑ & IDS↓ & HOTA↑ & DetA↑ & AssA↑ \\
\midrule
LSD \cite{von2012lsd} & 11.4 & 9.8 & / & / & / & / & / & / & / & /& /& / \\
Deepsort \cite{bewley2016simple} & 71.7 & 55.3 &211 & 56.4 & 56.9 & 55.8 & 72.9 & 57.8 &1947 & 58.7 & 60.1 &57.4 \\
Motdt \cite{chen2018real} & 69.1 & 54.0 &239 & 54.2 & 55.6 & 52.8 & 68.6 & 53.5 &2015 & 53.9 & 55.7 & 52.2 \\
CTracker \cite{peng2020chained} & 83.4 & 71.0 &240 & 72.9 & 74.3 & 71.5 & 69.3 & 61.7 &1660 & 58.2 & 63.9 & 53.0 \\
Sort+YoloX\cite{ge2021yolox} & 85.6 & 74.7 &\textcolor{red}{\textbf{130}} & 75.1 & 75.3 & 74.8 & 76.0 & 59.3 &1394 & 61.5 & 61.7 & 61.3 \\
Centertrack \cite{zhou2020tracking} & 86.3 & 77.1 & 174 & 81.0 &  \textcolor{blue}{\textbf{85.6}} & 75.9 & 72.5 & 64.8 & 1292 & 66.1 & \textcolor{blue}{\textbf{76.7}} & 56.9 \\
ByteTrack \cite{zhang2022bytetrack} & 91.0 & 83.3 &195 & 82.8 & 83.9 & 81.6 & 79.4 & 62.8 & \textcolor{blue}{\textbf{1211}} & 66.3 & 66.8 & \textcolor{blue}{\textbf{65.8}} \\
OCSORT \cite{cao2023observation} & \textcolor{blue}{\textbf{91.6}} &  \textcolor{blue}{\textbf{84.1}}  & \textcolor{blue}{\textbf{138}} &  \textcolor{blue}{\textbf{83.7}} & 84.9 &  \textcolor{blue}{\textbf{82.5}} & \textcolor{red}{\textbf{82.6}} &  \textcolor{blue}{\textbf{68.5}} & 1260 & \textcolor{blue}{\textbf{70.8}} & 73.3 & \textcolor{red}{\textbf{68.3}}\\
\rowcolor[gray]{0.9}
SDTNet (Ours) & \textcolor{red}{\textbf{91.8}} & \textcolor{red}{\textbf{87.7}} & 169 & \textcolor{red}{\textbf{87.8}} & \textcolor{red}{\textbf{90.9}} & \textcolor{red}{\textbf{84.8}} & \textcolor{blue}{\textbf{80.6}} & \textcolor{red}{\textbf{70.3}} & \textcolor{red}{\textbf{1070}} & \textcolor{red}{\textbf{73.6}} & \textcolor{red}{\textbf{81.6}}
& 65.6 \\
\bottomrule
\end{tabular}
}
\end{table*}

\subsection{Experimental Setup}

All experiments are performed using the mmdetection framework~\cite{chen2019mmdetection} with Pytorch \cite{paszke2019pytorch}. The SDT-Net model is trained on 8 NVIDIA 4090 GPUs. %For the SDTD dataset, DLA34 \cite{yu2018deep} is used as the backbone of the model. 
The entire model is trained for 60 epochs, and the initial learning rate is set to 3e-3,  which is decayed by 0.1 at the 20 epoch. The association radius $r$ to 200. Additionally, We adopt a downsampling strategy to generate paired images for training. Specifically, we first crop image patches of size 1524$\times$1524 from the original high-resolution images as ground truth. We employ the resolution of the input images is set to 1524$\times$1524 and the batch size is set to 2. Besides, the value of $\lambda_{\text{seg}}$ and $\lambda_{\text{emb}}$ are set to 1.0, $\lambda_{\text{hm}}$ is set to 10, and $\lambda_{\text{off}}$ is set to 0.1,

\subsection{Evaluation Metrics}
For evaluating the performance, we use standard MOT~\cite{bernardin2008evaluating} evaluation metrics. Specifically, we adopt commonly used metrics in the field, including Multi-Object Tracking Accuracy (MOTA), identity F1 score (IDF1), Identity Switches (IDs), and Higher Order Tracking Accuracy (HOTA)~\cite{luiten2021hota}. These metrics provide a comprehensive assessment of tracking performance. Notably, HOTA includes Detection Accuracy~(DetA) and Association Accuracy~(AssA), where DetA measures the accuracy of detecting objects regardless of their identities, and AssA evaluates how well identities are consistently maintained across frames.

\subsection{Comparison with Existing MOT Methods}

As shown in Tab.~\ref{tab3}, we compare our proposed SDT-Net with state-of-the-art MOT methods on the SDTD test set. Our method outperforms previous approaches in tracking space debris, particularly under dense debris conditions. 
% compared with OCSOR, SDTD achieves improvements of approximately 3.6\% in MOTA and 4.1\% in HOTA under Debris test set, and 1.8\% and 2.8\%, respectively, under multi-object conditions. These results highlight the effectiveness of our newly proposed SDTD. Additionally, Table~\ref{tab3} reports the detection results of other tracking methods for further comparison.
Especially, SDT-Net significantly outperforms CenterTrack in tracking performance. On the debris test set, it improves MOTA and HOTA by 10.6\% and 6.8\%, respectively, while on the dense debris test set, the improvements are 5.5\% and 7.5\%. These results demonstrate that SDT-Net achieves superior object association (higher AssA) and detection robustness (higher DetA). As shown in Tab.~\ref{tab3}, SDT-Net consistently outperforms CenterTrack across all key metrics, highlighting its effectiveness in complex space debris tracking scenarios.

\begin{figure}
\centering %表示居中
\includegraphics[width=0.8\columnwidth]{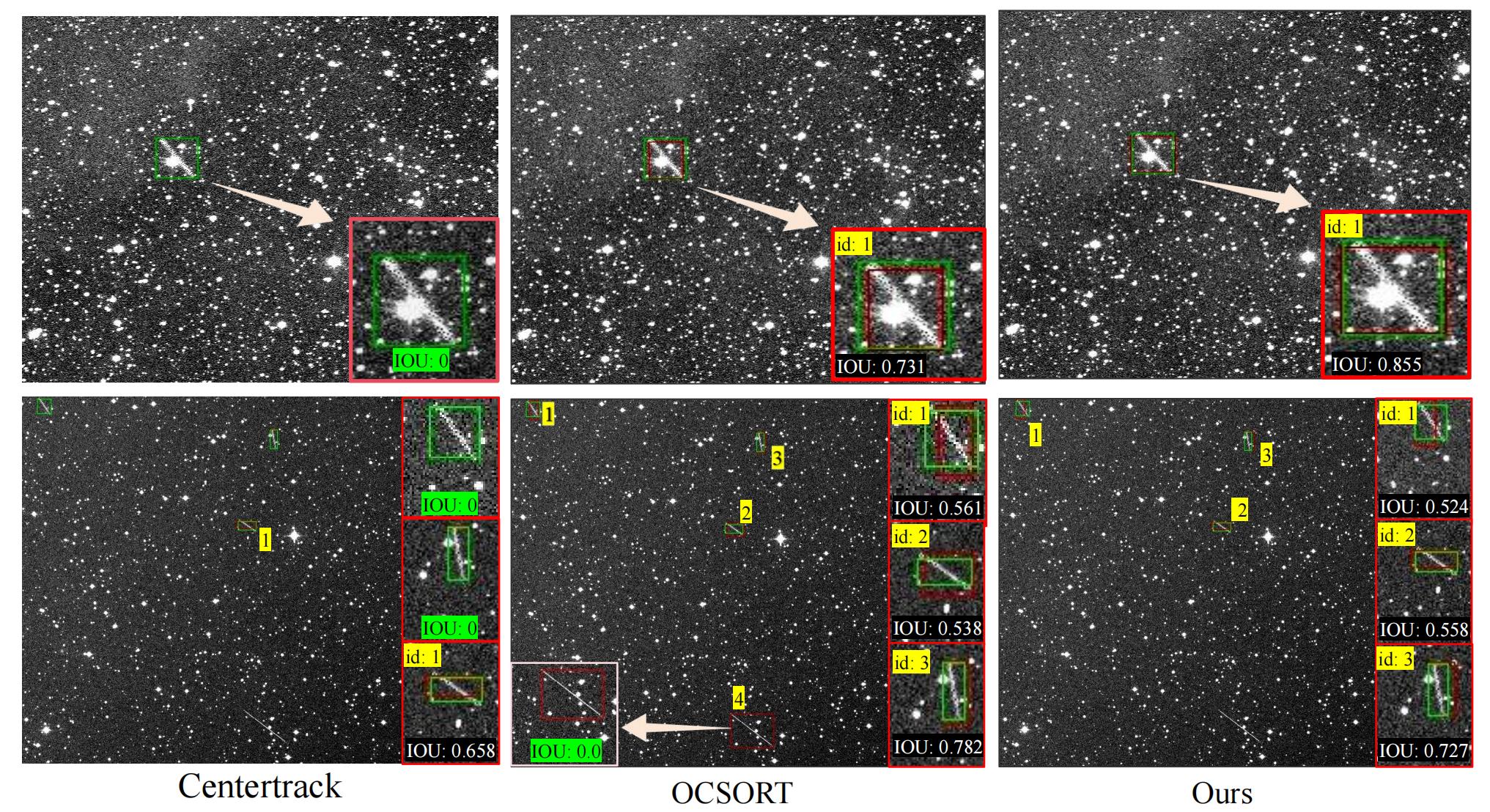}
\caption{Visualization example of the tracking results of the SDTD model on the test set. The green bounding box represents the GT, and the red bounding box represents the model prediction result. When the Intersection over Union~(IoU) between the predicted box and GT is 0, the presence of only a green box indicates a False Negative, while the presence of only a red box indicates a False Positive.}
\label{fig6}
\vspace{-8pt}
\end{figure}

\subsection{Ablation Study}

To analyze the contributions of different components in SDT-Net, we conduct an ablation study on the Debris Test and Dense Debris Test sets, as shown in Tab~\ref{tab2}. We evaluate three configurations: (a) Based on Centertrack, the center point and box size parameters are used to predict line source debris, and the debris offset module is also included; (b) LSE (Line Source Embedding), which models debris as line sources and directly locates their endpoint pairs; (c) RoI-FE (Region of Interest Feature Enhancement) further enhances the fragment feature representation and removes the debris offset module; and (d) Complete SDT-Net.

\textbf{Effect of Line Source Embedding~(LSE)}: Compared to CenterTrack (row a), introducing the line source embedding in SDT-Net (row b) leads to notable improvements across all metrics. Specifically, MOTA improves from 77.1\% to 85.5\% on the Debris Test set and from 64.8\% to 64.9\% on the Dense Debris Test set. This indicates that directly modeling debris as segments rather than center-based boxes provides better spatial localization, leading to improved object association (higher IDF1) and overall tracking performance.

\textbf{Region-of-Interest Feature Enhancement~(RoI-FE)}: Further incorporating segmentation masks (compare rows b and d) enhances the representation of debris features. This results in additional performance gains, with MOTA improving from 85.5\% to 87.7\% on the Debris Test set and from 64.9\% to 70.3\% on the Dense Debris Test set. Similarly, HOTA sees a significant increase, confirming that segmentation helps improve tracking robustness.

\textbf{Offset Module}: As shown in rows (c) and (d), introducing the debris offset module alone can significantly improve all indicators. Specifically, MOTA is improved from 86.0\% (row c) to 87.7\% on the fragmented test set, and from 62.9\% to 70.3\% on the dense fragmented test set. These improvements show that the offset module plays a key role in improving endpoint accuracy and overall tracking robustness. Finally, after integrating all modules, IDF1 is 80.6\%, MOTA is 70.3\%, and HOTA is 73.6\% on the dense fragmented test set.

% \begin{table}[htbp]
% \centering
% \caption{Ablation study of SDT-Net.}
% \label{tab4}
% % \small % 使用较小的字体
% \setlength{\tabcolsep}{2pt} % 进一步调整列间距
% \renewcommand{\arraystretch}{0.9} % 进一步调整行间距
% \begin{tabular}{c|ccc|ccc|ccc}
% \toprule
% &\multirow{2}{*}{LSE} & \multirow{2}{*}{RoI-FE} & \multirow{2}{*}{Offset} & \multicolumn{3}{c|}{Debris Test} & \multicolumn{3}{c}{Dense Debris Test} \\
% \cmidrule{4-9}
% & & & IDF1↑ & MOTA↑ & HOTA↑ & IDF1↑ & MOTA↑ & HOTA↑ \\
% \midrule
% (a)& & &\checkmark & 86.3 & 77.1 & 81.0 & 72.5 & 64.8 & 66.1 \\
% (b) &\checkmark &\checkmark & &  88.2 &86.0 &85.2 &73.5 &62.9 &63.0 \\
% (c) &\checkmark & &\checkmark  & 90.4 & 85.5 & 84.8 & 78.5 & 64.9 & 65.3 \\
%  % & \checkmark & & 91.1 & 87.2 & 87.1 & 76.1 & 71.6 & 70.3 \\
% \rowcolor[gray]{0.9}
% (d)&\checkmark & \checkmark &\checkmark & \textbf{91.8} & \textbf{87.7} & \textbf{87.8} & \textbf{80.6} & \textbf{70.3} & \textbf{73.6} \\
% \bottomrule
% \end{tabular}
% \end{table}

\begin{table}[htbp]
\centering
\caption{Ablation study of SDT-Net.}
\label{tab2}
\setlength{\tabcolsep}{5pt} % 适当增加列间距
\renewcommand{\arraystretch}{1.2} % 适当增加行间距
\begin{tabular}{c|ccc|cccc|cccc}
\toprule
& \multirow{2}{*}{LSE} & \multirow{2}{*}{RoI-FE} & \multirow{2}{*}{Offset Module} & \multicolumn{4}{c|}{Debris Test~( $n \boldsymbol{\leq} 2$)} & \multicolumn{4}{c}{Dense Debris Test~($1 \boldsymbol{\leq} n \boldsymbol{\leq} 5$)} \\
\cmidrule{5-12}
& & & & IDF1$\uparrow$ & MOTA$\uparrow$ & HOTA$\uparrow$  & DetA$\uparrow$
& IDF1$\uparrow$ & MOTA$\uparrow$ & HOTA$\uparrow$ & DetA$\uparrow$ \\
\midrule
(a) & & & \checkmark & 86.3 & 77.1 & 81.0 &85.6 & 72.5 & 64.8 & 66.1 &76.7 \\
(b) & \checkmark & & \checkmark & 90.4 & 85.5 & 84.8 & 88.3 & 78.5 & 64.9 & 65.3 & 79.1\\
(c) & \checkmark & \checkmark & & 88.2 & 86.0 & 85.2 & 88.9 & 73.5 & 62.9 & 63.0  & 78.6 \\
\rowcolor[gray]{0.9}
(d) & \checkmark & \checkmark & \checkmark & \textbf{91.8} & \textbf{87.7} & \textbf{87.8} & \textbf{90.9} & \textbf{80.6} & \textbf{70.3} & \textbf{73.6} & \textbf{81.6}\\
\bottomrule
\end{tabular}
\end{table}

The complete SDT-Net (row c) achieves the best performance, demonstrating that combining LSE and RoI-FE leads to the most accurate debris tracking. These results highlight that directly predicting line source endpoints improves accuracy, while segmentation masks further strengthen debris features, particularly in challenging dense debris scenarios.

\begin{figure}
\centering %表示居中
\includegraphics[width=1.0\columnwidth]{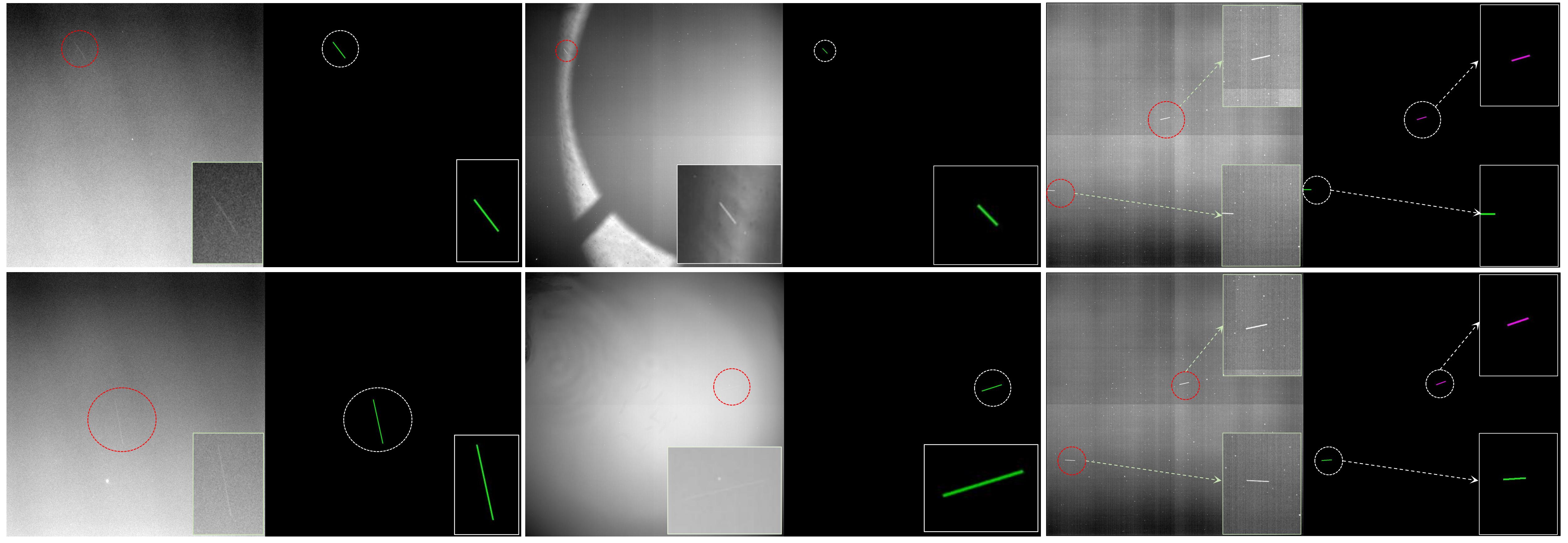}
\caption{Visualization of sample tracking results of real-world observations. Left: real-world debris observation. Right: model predicts tracking results, represented by predicted endpoints and connecting segments forming the trajectory (superimposed on a blank background). Different colors correspond to different debris IDs.}
\label{fig11}
\vspace{-8pt}
\end{figure}

\subsection{Qualitative Results}

We compare it with state-of-the-art tracking methods, including OCSORT and CenterTrack, in complex scenarios. Fig.~\ref{fig6} illustrates the tracking results on datasets containing space debris with cluttered backgrounds, occlusions, and varying object densities.

CenterTrack fails to maintain debris tracking when occluded by light sources, whereas SDTD successfully detects and tracks the debris in such situations. In scenarios with dense debris, OCSORT produces false positives, while SDTD exhibits superior spatial detection and tracking capabilities. This reflects its ability to accurately associate and locate dense debris. 

To further verify the effectiveness of SDT-Net, we evaluate the model with real-world debris data. Due to the scarcity of real-world debris observation data, we finally select 36 video sequences containing debris (a total of 2,228 frames) and invited astronomy experts to annotate the real-world observation data. As shown in Table~\ref{tab4}, our method (SDT-Net) achieves a MOTA score exceeding 73\%, demonstrating strong real-world performance. Fig~\ref{fig11} further illustrates the effectiveness of our approach. The left side of the figure presents raw debris observations, while the right side visualizes the predicted tracking results. Notably, the right side does not represent ground truth but rather our model’s predictions. For clarity, we display each tracked object using a line connecting its two key points instead of conventional bounding boxes. In practical applications, these tracked trajectories are overlaid on blank background images to facilitate interpretation. Each color in the visualization corresponds to a distinct debris ID, reflecting the model’s ability to distinguish and track multiple objects.

\begin{table}[h]
\centering
\caption{Evaluation of SDT-Net on the Antarctic Station real-world data.}
\label{tab4}
\setlength{\tabcolsep}{8pt} % 调整列间距
\renewcommand{\arraystretch}{1.2} % 调整行间距
\begin{tabular}{lccc}
\toprule
Method & MOTA$\uparrow$ & HOTA$\uparrow$ & DetA$\uparrow$  \\
\midrule
Deepsort~\cite{bewley2016simple} &48.7 &48.9 &51.8  \\
Motdt~\cite{chen2018real} & 47.4 &45.2 &47.6  \\
CTracker~\cite{peng2020chained} &51.9 &45.8 &54.5   \\
Sort+YoloX~\cite{ge2021yolox} &57.4 &57.4 &58.9  \\
CenterTrack~\cite{zhou2020tracking} & 64.7 & 58.3 & 68.7 \\
ByteTrack~\cite{zhang2022bytetrack} &66.8 &62.4 &70.1   \\
OCSORT~\cite{cao2023observation}      & 69.3 & 66.8 & 72.2 \\
\rowcolor[gray]{0.9}
SDT-Net     & \textbf{73.2} & \textbf{68.1} & \textbf{75.8} \\
\bottomrule
\end{tabular}
\end{table}

\section{Conclusion}

In this paper, we introduce SDTD, the first benchmark dataset for space debris tracking. SDTD consisting of 18,040 videos from ZTF with 62,562 frames and 250,000 synthetic debris. SDTD provides diverse scenarios, including complex backgrounds and dense debris. Based on this, we propose a deep learning-based tracking network, SDTD, which achieves state-of-the-art performance. 
Furthermore, we conduct a user study on real-world data to assess the model’s generalization ability. The results demonstrate that SDT-Net not only performs well on real debris scenes (e.g., Antarctic Station data) but also exhibits transferability to out-of-distribution scenarios, ensuring robust tracking capabilities in challenging environments.

\bibliographystyle{unsrt}  
\bibliography{references}  %%% Remove comment to use the external .bib file (using bibtex).
%%% and comment out the ``thebibliography'' section.

%%% Comment out this section when you \bibliography{references} is enabled.
% \begin{thebibliography}{1}

% \bibitem{kour2014real}
% George Kour and Raid Saabne.
% \newblock Real-time segmentation of on-line handwritten arabic script.
% \newblock In {\em Frontiers in Handwriting Recognition (ICFHR), 2014 14th
%   International Conference on}, pages 417--422. IEEE, 2014.

% \bibitem{kour2014fast}
% George Kour and Raid Saabne.
% \newblock Fast classification of handwritten on-line arabic characters.
% \newblock In {\em Soft Computing and Pattern Recognition (SoCPaR), 2014 6th
%   International Conference of}, pages 312--318. IEEE, 2014.

% \bibitem{hadash2018estimate}
% Guy Hadash, Einat Kermany, Boaz Carmeli, Ofer Lavi, George Kour, and Alon
%   Jacovi.
% \newblock Estimate and replace: A novel approach to integrating deep neural
%   networks with existing applications.
% \newblock {\em arXiv preprint arXiv:1804.09028}, 2018.

% \end{thebibliography}
\clearpage
% Supplementary Material

% \subsection{Complex Skylight Backgrounds}

\begin{figure*}
\centering %表示居中
\includegraphics[width=1.0\columnwidth]{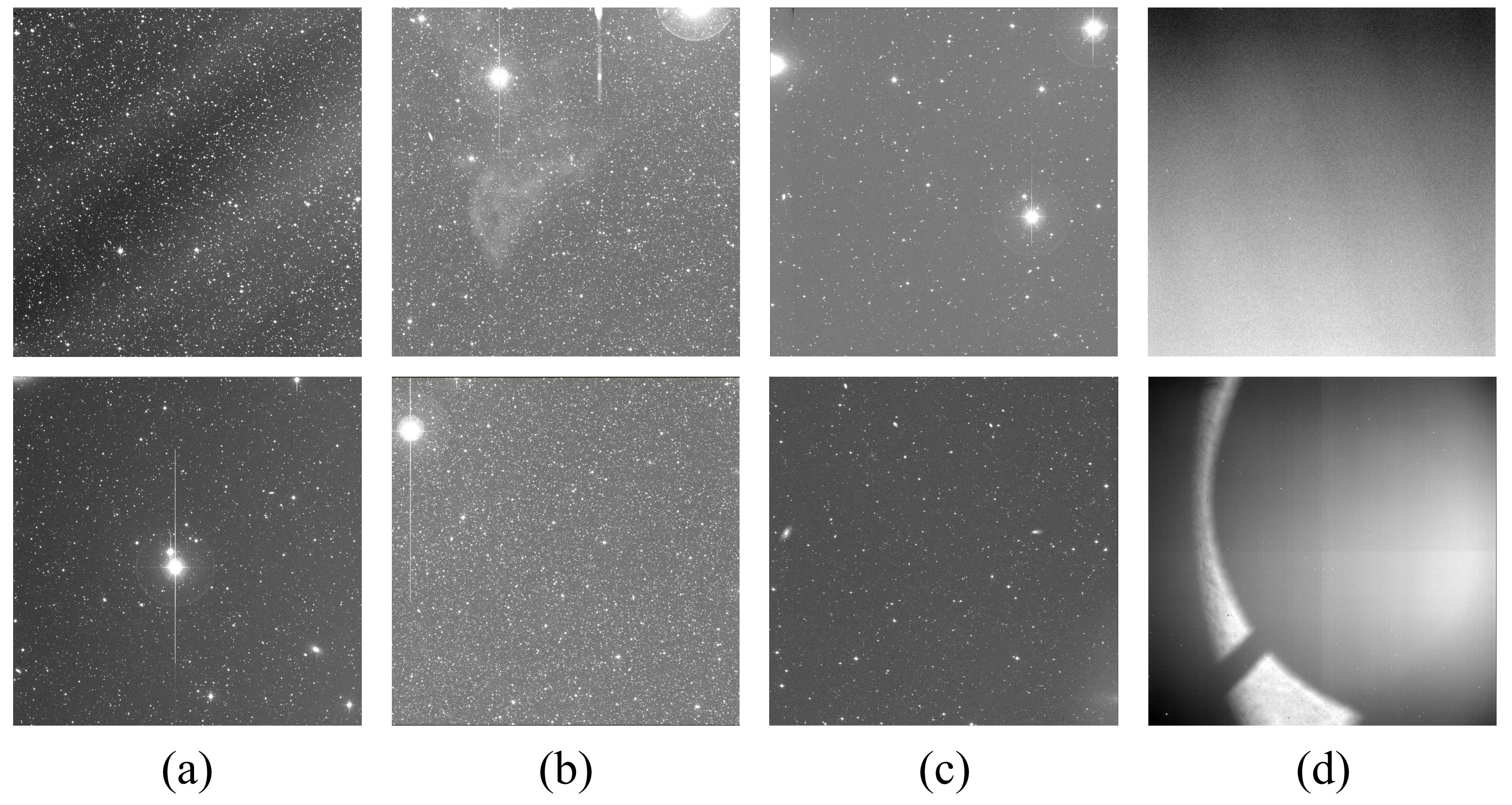}
\caption{Data example of complex skylight background.}
%图片的名称
\label{fig7}
%图片的标签，用于文章中的引用，注意到标签的数字与实际文章显示的数字可能不同
\end{figure*}

\begin{figure}
\centering %表示居中
\includegraphics[width=1.0\columnwidth]{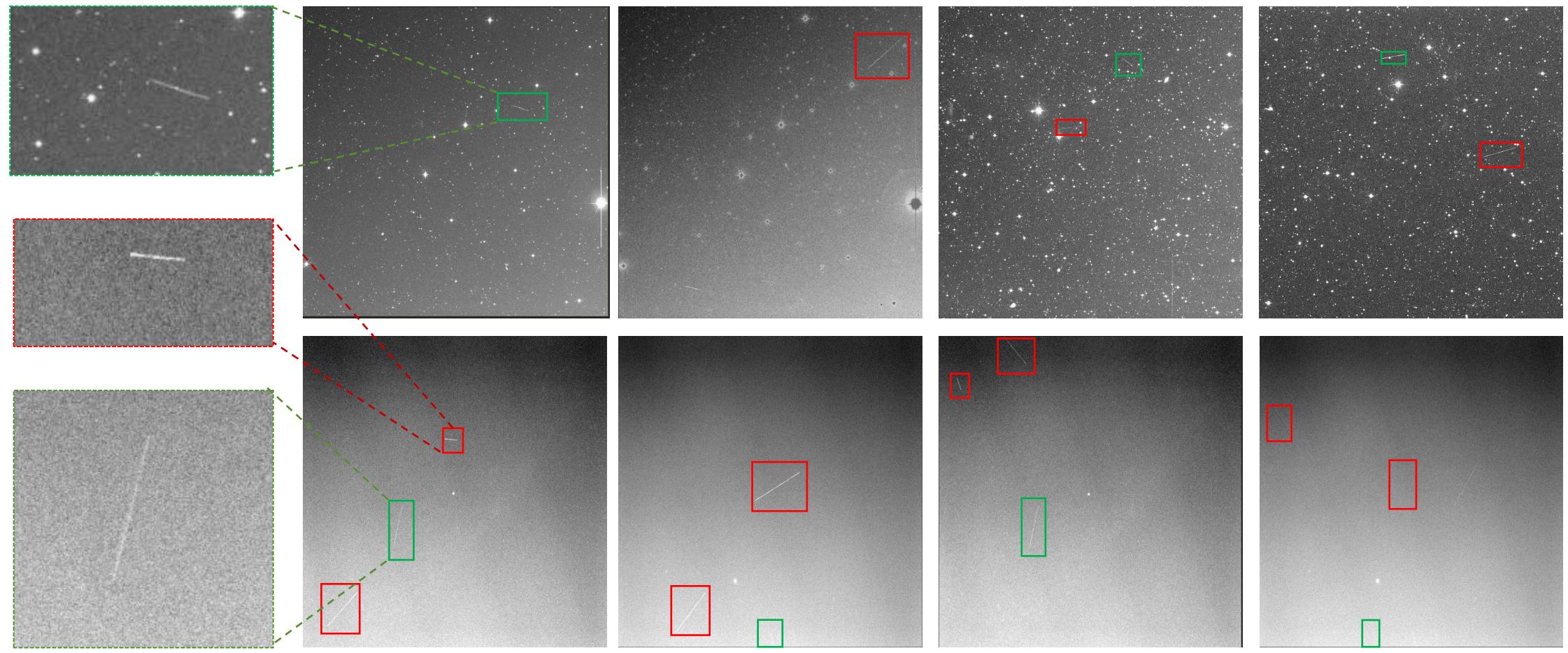}
\caption{Comparison of debris simulation and real-world debris. The green bounding box represents real-world debris, and the red bounding box represents simulated debris.}
%图片的名称
\label{fig2}
%图片的标签，用于文章中的引用，注意到标签的数字与实际文章显示的数字可能不同
\end{figure}

\begin{figure*}
\centering %表示居中
\includegraphics[width=0.5\columnwidth]{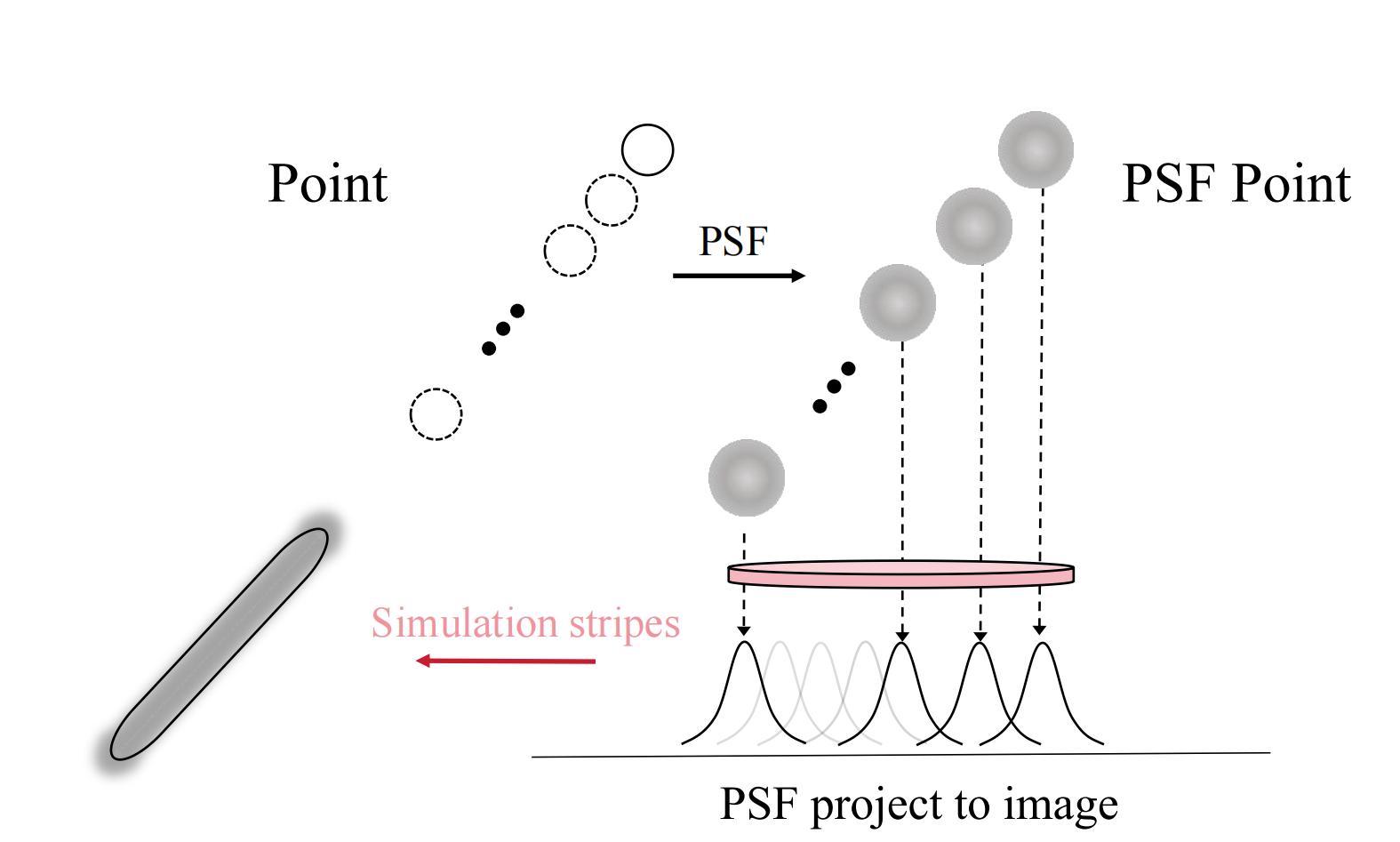}
\caption{Dynamic imaging process of space debris.}
%图片的名称
\label{fig2}
%图片的标签，用于文章中的引用，注意到标签的数字与实际文章显示的数字可能不同
\end{figure*}

\begin{figure*}
\centering %表示居中
\includegraphics[width=1.0\columnwidth]{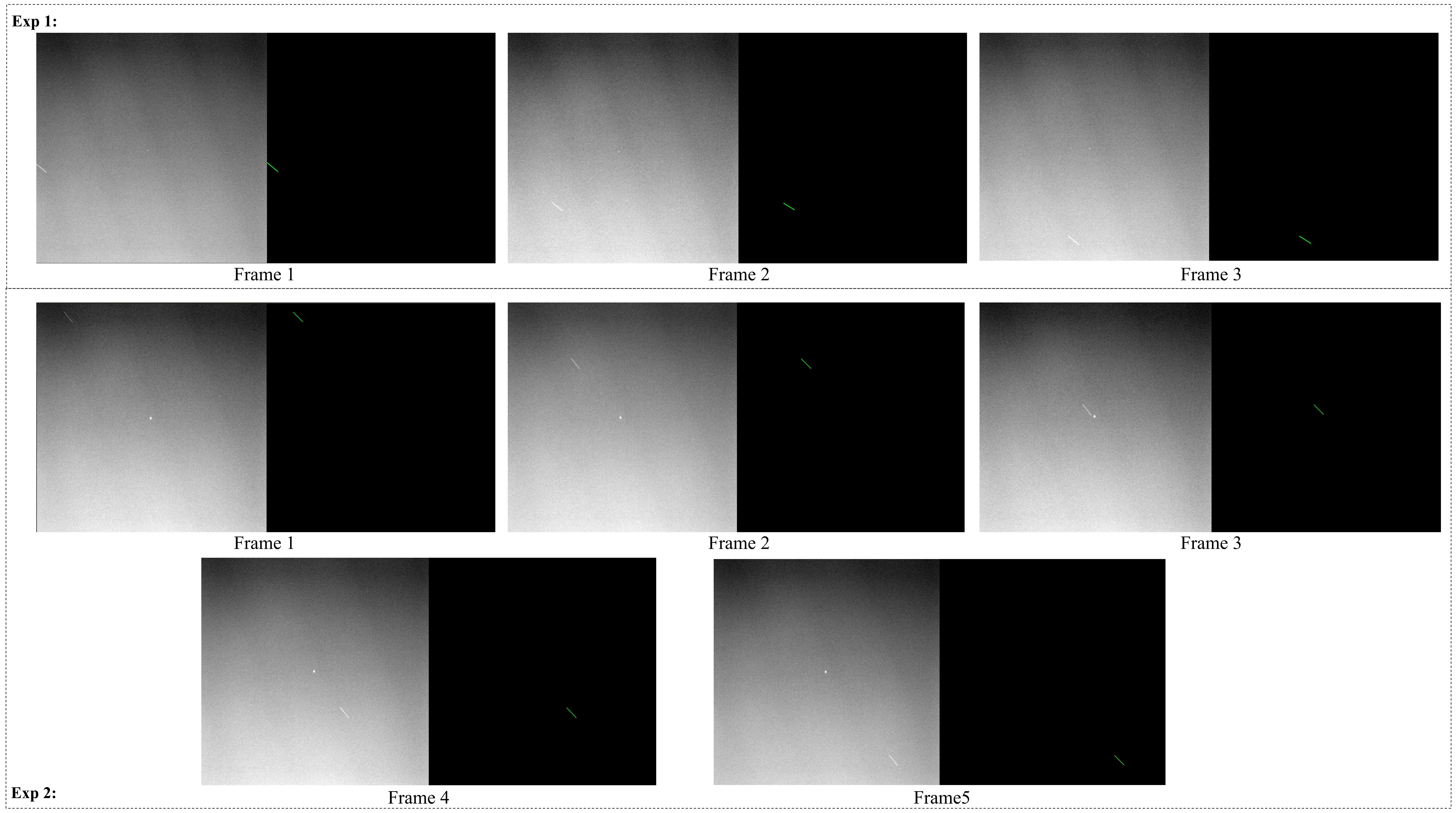}
\caption{SDT-Net tracking results in the real-world data.}
%图片的名称
\label{fig2}
%图片的标签，用于文章中的引用，注意到标签的数字与实际文章显示的数字可能不同
\end{figure*}

\begin{figure*}
\centering %表示居中
\includegraphics[width=1.0\columnwidth]{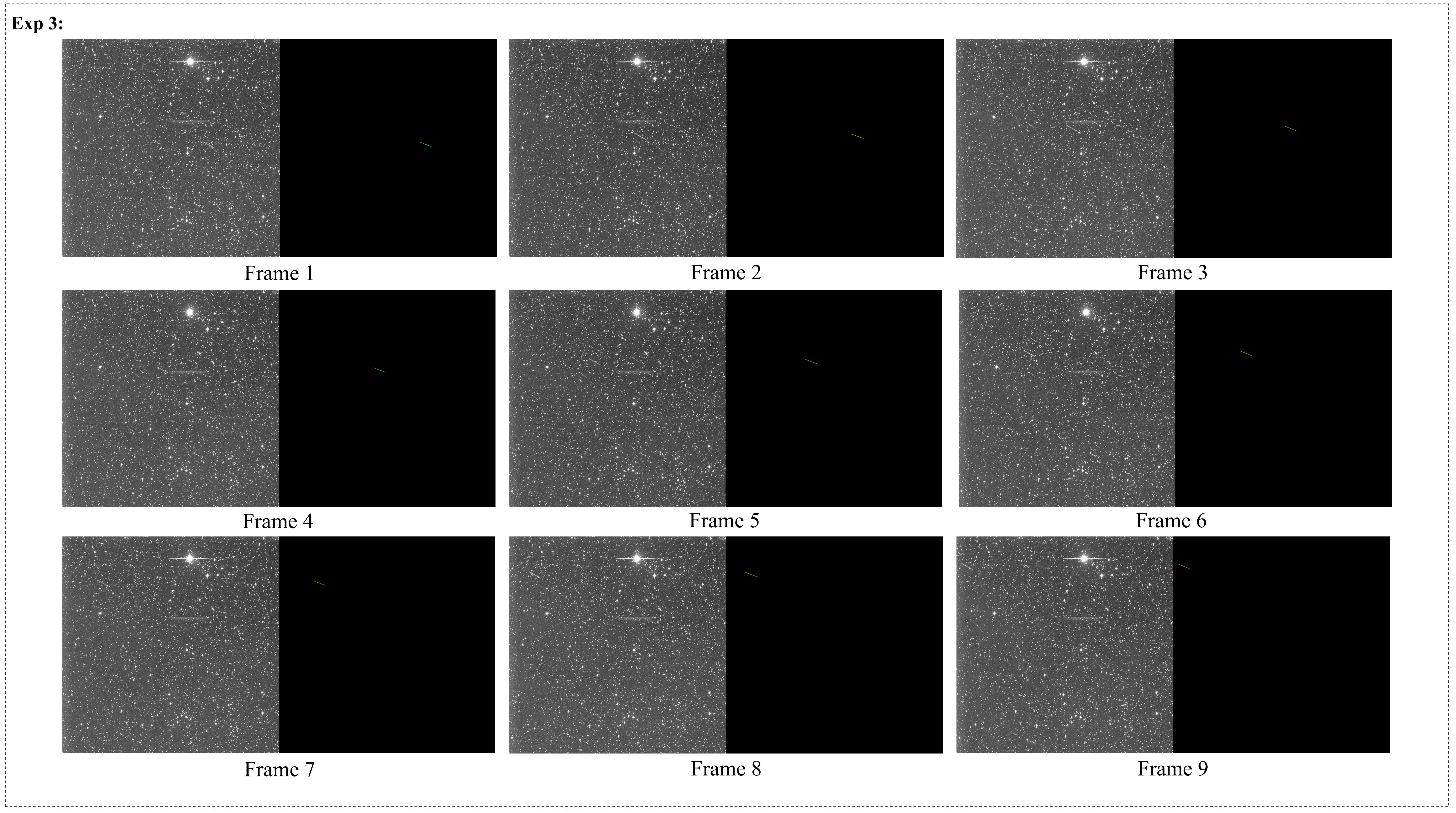}
\caption{SDT-Net test tracking result example in SDTD.}
%图片的名称
\label{fig2}
%图片的标签，用于文章中的引用，注意到标签的数字与实际文章显示的数字可能不同
\end{figure*}

\begin{figure*}
\centering %表示居中
\includegraphics[width=1.0\columnwidth]{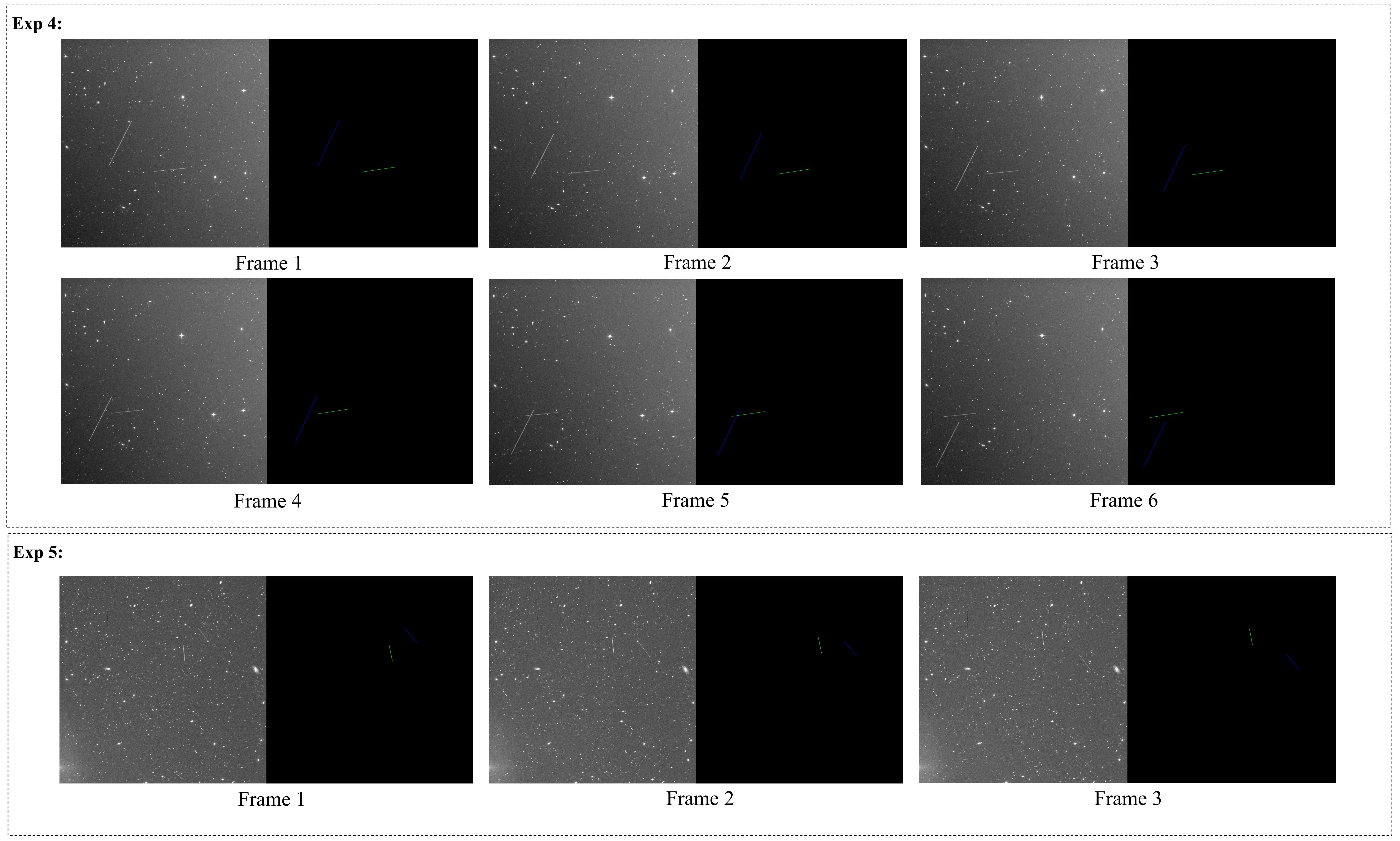}
\caption{SDT-Net test tracking result example in SDTD, including at least two debris.}
%图片的名称
\label{fig2}
%图片的标签，用于文章中的引用，注意到标签的数字与实际文章显示的数字可能不同
\end{figure*}

\end{document}